\documentclass[10pt,twocolumn,letterpaper]{article}

\usepackage{cvpr}              

\usepackage{graphicx}
\usepackage{amsmath}
\usepackage{amssymb}
\usepackage{booktabs}
\usepackage{multirow}

\usepackage[pagebackref,breaklinks,colorlinks]{hyperref}

\usepackage[capitalize]{cleveref}
\crefname{section}{Sec.}{Secs.}
\Crefname{section}{Section}{Sections}
\Crefname{table}{Table}{Tables}
\crefname{table}{Tab.}{Tabs.}

\newcommand*\rot{\rotatebox{30}}

\def\cvprPaperID{*****} 
\def\confName{CVPR}
\def\confYear{2023}

\begin{document}

\title{The Second-place Solution for CVPR VISION 23 Challenge Track 1 - Data Effificient Defect Detection}

\author{Xian Tao\thanks{Equal contribution.}~~$^{1,2}$ ~~~~ Zhen Qu\footnotemark[1]~~$^{2}$ ~~~~ Hengliang Luo\footnotemark[1]~~$^{1}$ ~~~~ Jianwen Han$^1$  \\[-2pt]
Yonghao He$^{1}$ ~~ Danfeng Liu$^{1}$ ~~ Chengkan Lv$^{1,2}$ ~~ Fei Shen$^{1,2}$ ~~ Zhengtao Zhang$^{1,2}$\\
$^1$CASI Vision Technology Co., Ltd., Luoyang, China \\
$^2$CAS Engineering Laboratory for Intelligent Industrial Vision, \\[-2pt]
Institute of Automation, Chinese Academy of Sciences, Beijing, China \\
{\tt\small taoxian2013@ia.ac.cn~~quzhen22@mails.ucas.ac.cn~~luo@hengliang.me~~2641410938@qq.com} \\[-2pt]
{\tt\small \{yonghao.he,~danfeng.liu,~chengkan.lv\}@casivision.com~~\{fei.shen,~zhengtao.zhang\}@ia.ac.cn}}

\maketitle

\begin{abstract}
   The Vision Challenge Track 1 for Data-Effificient Defect Detection requires competitors to instance segment 14 industrial inspection datasets in a data-defificient setting. This report introduces the technical details of the team \textbf{Aoi-overfifitting-Team} for this challenge. Our method focuses on the key problem of segmentation quality of defect masks in scenarios with limited training samples. Based on the Hybrid Task Cascade (HTC) instance segmentation algorithm, we connect the transformer backbone (Swin-B) through composite connections inspired by CBNetv2 to enhance the baseline results. Additionally, we propose two model ensemble methods to further enhance the segmentation effect: one incorporates semantic segmentation into instance segmentation, while the other employs multi-instance segmentation fusion algorithms. Finally, using multi-scale training and test-time augmentation (TTA), we achieve an average mAP@0.50:0.95 of more than 48.49\% and an average mAR@0.50:0.95 of 66.71\% on the test set of the Data Effificient Defect Detection Challenge. The code is available at \url{https://github.com/love6tao/Aoi-overfitting-team}
\end{abstract}

\section{Introduction}
\label{sec:intro}

The use of deep learning for visual inspection has become increasingly prevalent in the realm of industrial defect detection due to its efficiency and unparalleled accuracy. This includes tasks such as unmanned aerial vehicle (UAV) patrol inspection of power equipment~\cite{tao2018detection}, detecting weak scratches on industrial surfaces~\cite{tao2020industrial}, identifying copper wire defects in deep hole parts~\cite{tao2018wire}, and detecting conductive particles on chip and glass surfaces~\cite{tao2021conductive}, among others. However, obtaining labeled defect data in industrial manufacturing scenarios is difficult, expensive, and time consuming, making visual-based industrial inspection more challenging. To address this issue, the CVPR VISION 23 Challenge Track 1 - Data Efficient Defect Detection competition is launched~\cite{vision_datasets_2023}. 

\begin{figure}[t]
  \centering
  \includegraphics[width=0.98\linewidth]{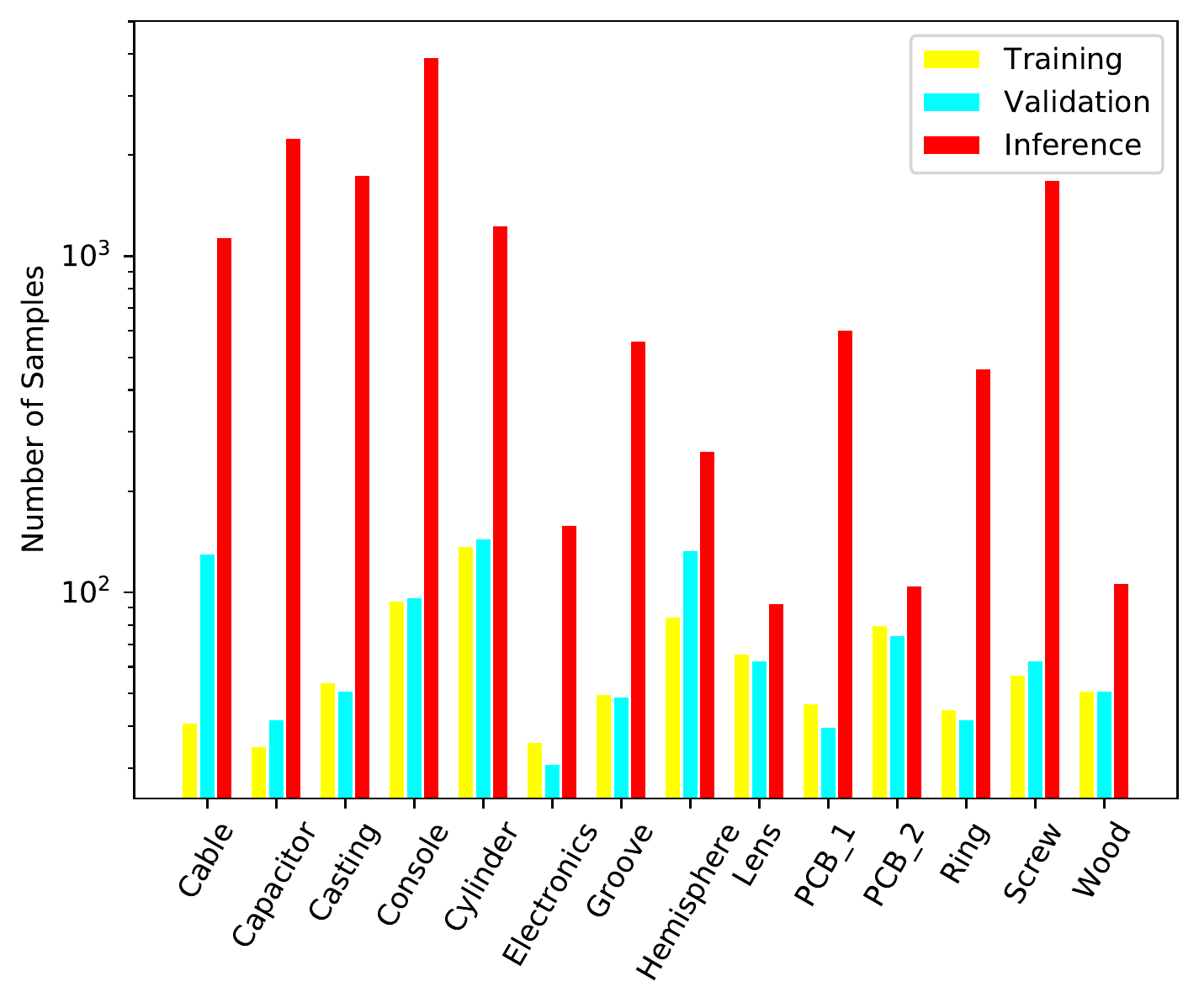}
  \caption{Number of training, validation, and inference samples for each dataset in this challenge. Note that the number of training and validation samples is significantly smaller than the number of inference samples, so the Y-Axis is in log scale for clarity.}
  \label{fig:dataset_num}
\end{figure}

The competition dataset consists of 14 defect datasets from real-world scenes, with the most remarkable feature being that the number of test samples far exceeds the number of training samples. Some datasets, such as the capacitor and electronics datasets, contain no more than 40 training samples. \cref{fig:dataset_num} shows sample number of each dataset. Moreover, certain images in the datasets exhibit significant-scale changes. The majority of boxes encompass only 10\% of the image, whereas some boxes can cover the entire image. Furthermore, there are significant differences in the background and shape of defect textures across the 14 
datasets, making it a huge challenge to construct a unified algorithm framework that yields satisfactory results on each dataset. To tackle these problems, we train a strong baseline model with Swin Transformer~\cite{liu2021swin} and CBNetV2~\cite{liang2107cbnetv2} Backbone, then two model ensemble methods are employed to further promote the segmentation performance.We will introduce our pipeline and detailed components in \cref{sec:app}. The experiment results and ablation studies are shown in \cref{sec:exp}.

\section{Approach}
\label{sec:app}

\begin{figure}[t]
  \centering
  \includegraphics[width=0.98\linewidth]{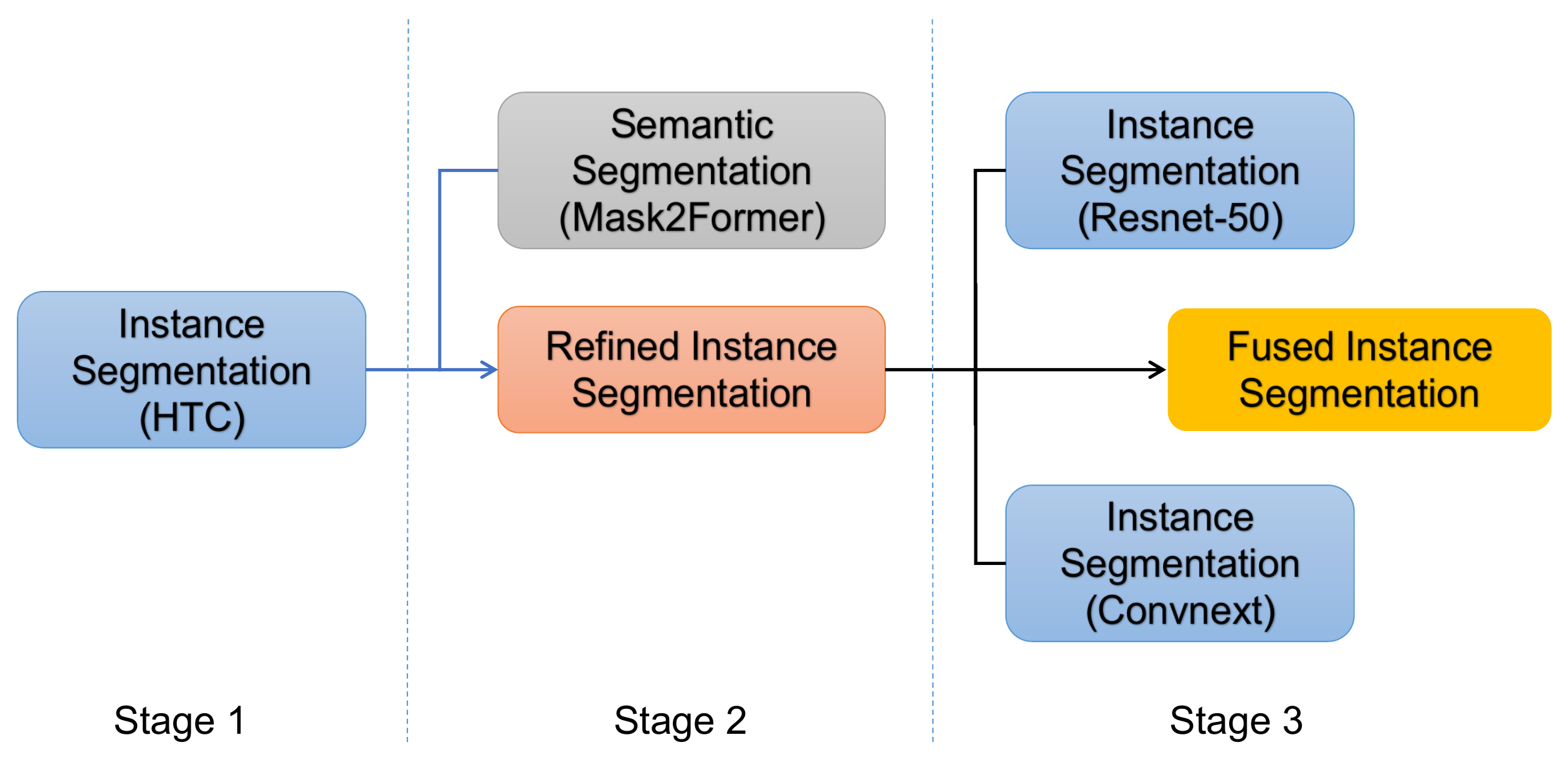}
  \caption{Pipeline of our method. It includes three stages: 1) training a powerful instance segmentation model as a baseline, 2) using semantic segmentation to refine the instance segmentation, and 3) fusing three instance segmentation models to obtain the final result.}
  \label{fig:pipeline}
\end{figure}

In this section, an effective pipeline comprising three parts is presented. A powerful single instance segmentation model is first trained as baseline using Hybrid Task Cascade, with Swin Transformer and CBNetV2 as its backbones, as shown in \cref{fig:pipeline}. Following this, a robust semantic segmentation model is trained using Mask2Former~\cite{cheng2022masked} to further refine segmentation performance, with the results of semantic segmentation fused with those of instance segmentation. Lastly, three instance segmentation outcomes are fused to further improve the segmentation effect for final submission.

\subsection{Base instance segmentation model}

\begin{figure}[t]
  \centering
  \includegraphics[width=0.98\linewidth]{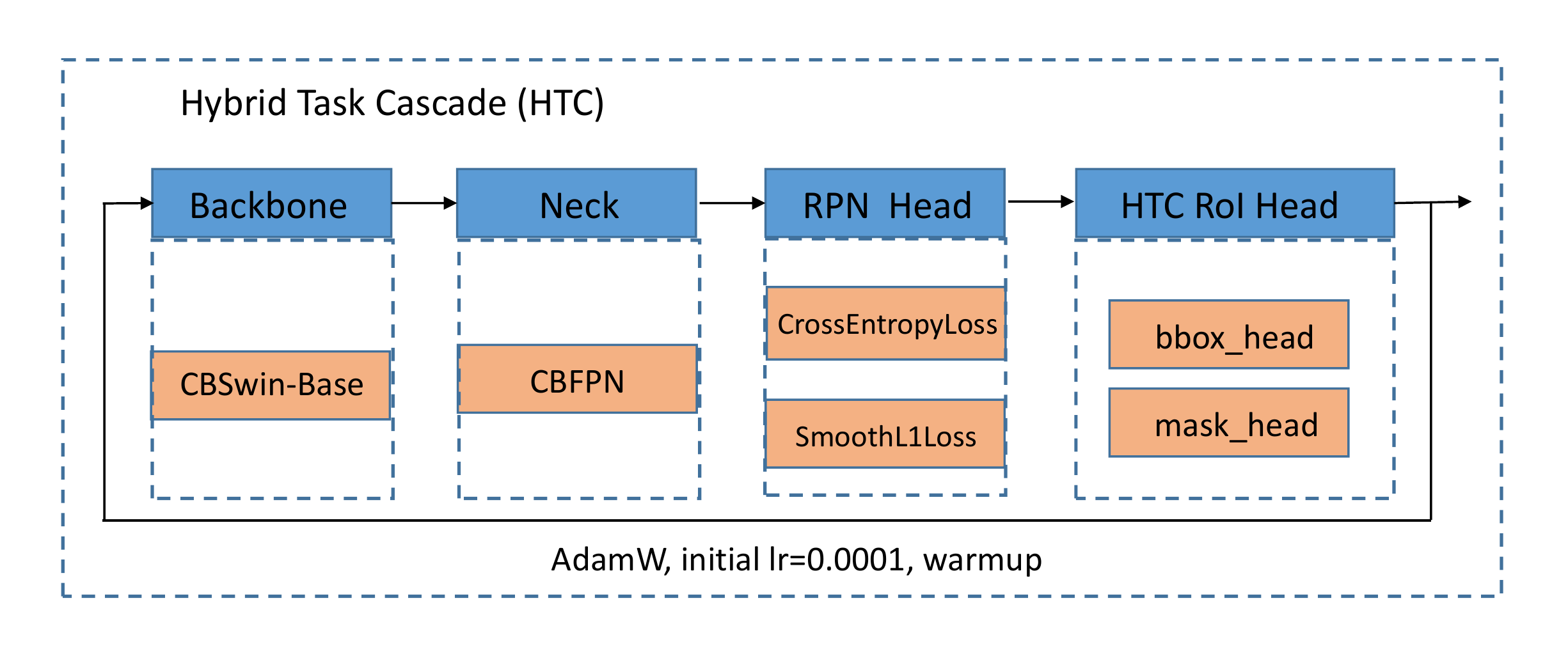}
  \caption{The architecture of our baseline model.}
  \label{fig:htc}
\end{figure}

Our base instance segmentation model is built upon the Hybrid Task Cascade (HTC)~\cite{chen2019hybrid} detector, which utilizes the CBSwinBase backbone with the CBFPN~\cite{liang2107cbnetv2} architecture. HTC is a robust cascade architecture for instance segmentation tasks that cleverly blends detection and segmentation branches for joint multi-stage processing, progressively extracting more discriminative features at each stage. To avoid the need for extra stuff segmentation annotations, we removed the semantic head from our solution. The recent advancements in the Vision Transformer have been remarkable for various visual tasks, and therefore, we adopt the Swin Transformer as our backbone. The Swin Transformer introduces an efficient window attention module in a hierarchical feature architecture, with a linear computational complexity concerning the input image size. For our work, we employ the Swin-B network as our basic backbone, pre-trained on the Imagenet-22k dataset. To further enhance performance, we are inspired by the CBNetv2 algorithm and group two identical Swin-B networks through composite connections. The model architecture's key components are illustrated in \cref{fig:htc}.

\subsection{Incorporating semantic segmentation into instance segmentation}

Although a single model can achieve good segmentation results, the results of instance segmentation are often incomplete, especially when the IoU threshold is set too high, which may have a negative impact on the mask mAP~\cite{chen2019hybrid}. Therefore, we use the output of the semantic segmentation model to supplement the results of the instance segmentation model.

Our semantic segmentation model is based on Mask2Former~\cite{cheng2022masked}, with Swin-L as the backbone, and the input image size of its network is 512 $\times$ 512. The pre-trained weight comes from the ADE20K dataset~\cite{zhou2019semantic}. In order to train the semantic segmentation network, we convert multi-defect labels into binary labels representing the background and defects. The final semantic segmentation of 14 datasets are shown in \cref{tab:seg_result}.

\begin{figure}[t]
  \centering
  \includegraphics[width=0.98\linewidth]{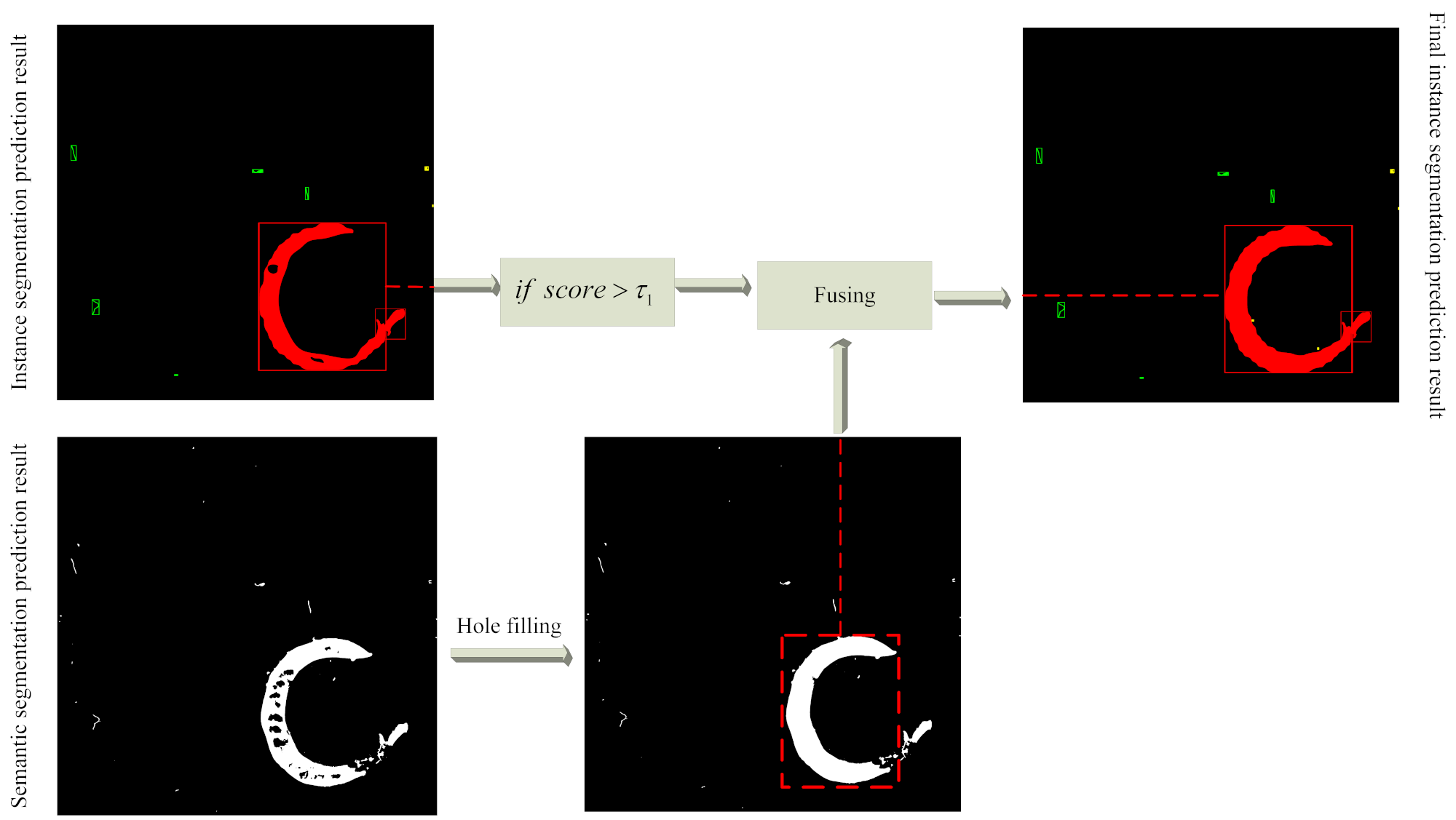}
  \caption{The process of incorporating semantic segmentation into instance segmentation.}
  \label{fig:fuse_semantic}
\end{figure}

For the fusion strategy, we combine the instance segmentation results with the semantic segmentation results at the same pixel positions, producing a new instance segmentation result, shown in \cref{fig:fuse_semantic}. Since the semantic segmentation task divides pixels into two categories, defect and back ground, the predicted class of the bounding box (bbox) in the instance segmentation task determines the actual class of pixels. It is important to note that only predicted instances with a bbox confidence greater than a threshold value of $\tau_1$ are considered for fusion with semantic segmentation results. In the competition, we set $\tau_1$ to 0.5 to achieve the best segmentation performance. 

\begin{table*}[t]
  \centering
  \begin{tabular}{p{1.5cm} p{0.7cm} p{0.7cm} p{0.7cm} p{0.7cm} p{0.7cm} p{0.7cm} p{0.7cm} p{0.7cm} p{0.7cm} p{0.7cm} p{0.7cm} p{0.7cm} p{0.7cm} p{0.7cm} }
    \toprule
    Dataset & \rot{Cable} & \rot{Capacitor} & \rot{Casting} & \rot{Console} & \rot{Cylinder} & \rot{Electronics} & \rot{Groove} & \rot{Hemisphere} & \rot{Lens} & \rot{PCB\_1} & \rot{PCB\_2} & \rot{Ring} & \rot{Screw} & \rot{Wood} \\
    \midrule
    mean IoU & 47.00 & 60.38 & 54.29 & 72.26 & 97.76 & 60.66 & 25.66 & 48.00 & 65.88 & 75.71 & 92.41 & 55.00 & 73.64 & 39.00\\
    \bottomrule
  \end{tabular}
  \caption{Semantic segmentation results obtained using Mask2Former. The results presented here were obtained from the validation dataset.}
  \label{tab:seg_result}
\end{table*}

\subsection{Fusion of multiple instance segmentations}

\begin{figure}[t]
  \centering
  \includegraphics[width=0.98\linewidth]{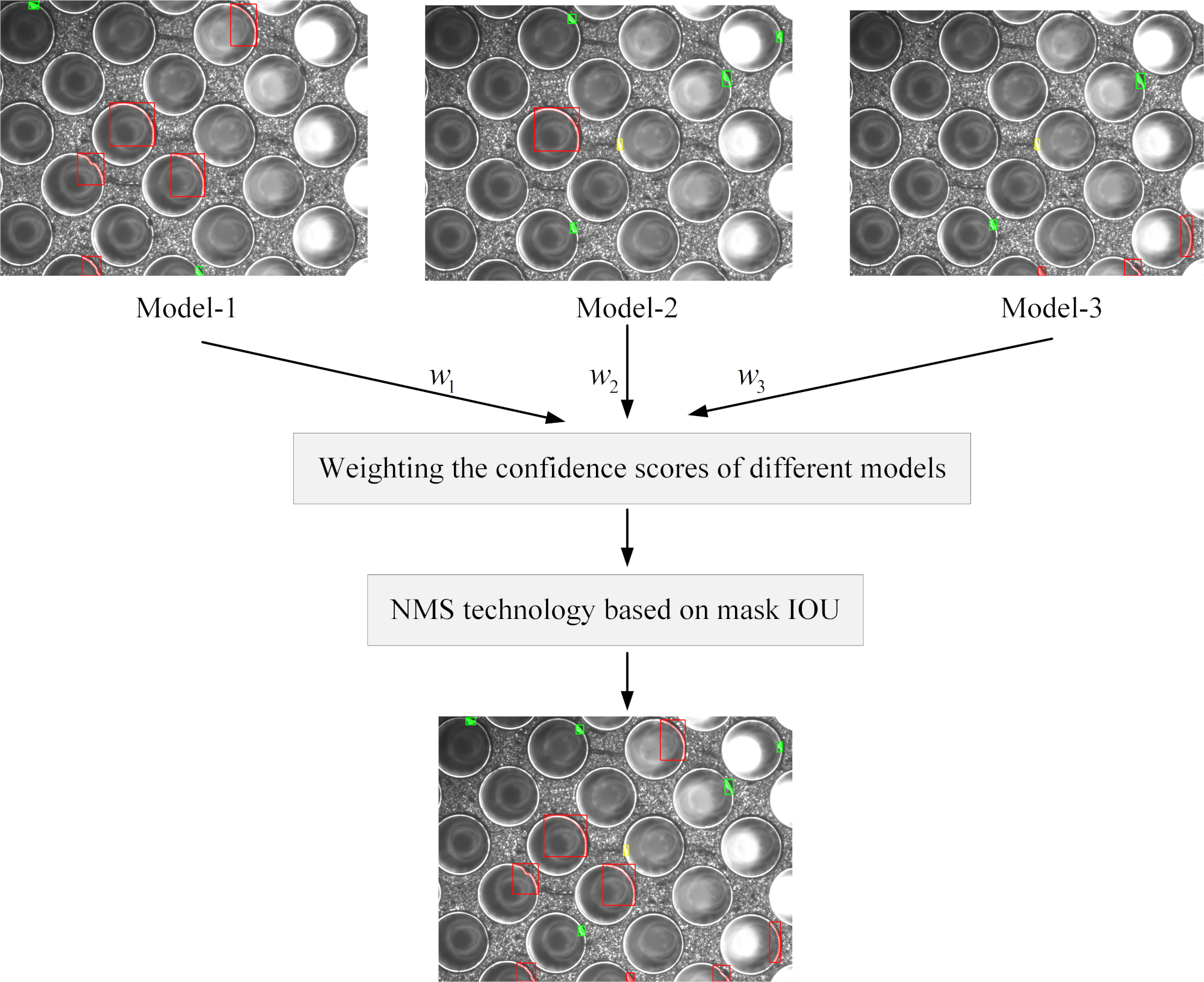}
  \caption{The process of fusion strategy.}
  \label{fig:fuse_multiple}
\end{figure}

Our experiments have revealed that different instance segmentation backbones can produce complementary results. This means that fusing the instance segmentation results of different backbones can lead to better recall for the model. However, improved recall often comes at the cost of decreased detection precision. To address this issue, we have designed a fusion strategy, as illustrated in \cref{fig:fuse_multiple}. 

In our experiments, we refer to model-1, model-2, and model-3 as the HTC, Cascade Mask rcnn-ResNet50, and Cascade Mask rcnn-ConvNext models, respectively. Specifically, the greater the mask-based mAP, the higher the weight given to the model to minimize the loss of precision. Finally, all confidence adjusted model detections undergo further processing using mask IoU-based NMS technology to filter out redundant detections. Our proposed multi-model ensemble technique has been shown in experiments to achieve a stable increase in mAR.

\section{Experiments}
\label{sec:exp}

\subsection{Datasets}

\begin{figure*}
  \centering
  \begin{subfigure}{0.138\linewidth}
    \includegraphics[width=0.98\linewidth]{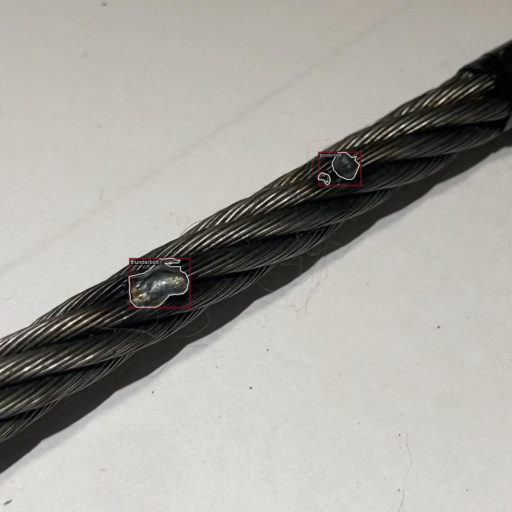}
    \caption*{(1)~Cable}
  \end{subfigure}
  \hfill
  \begin{subfigure}{0.138\linewidth}
    \includegraphics[width=0.98\linewidth]{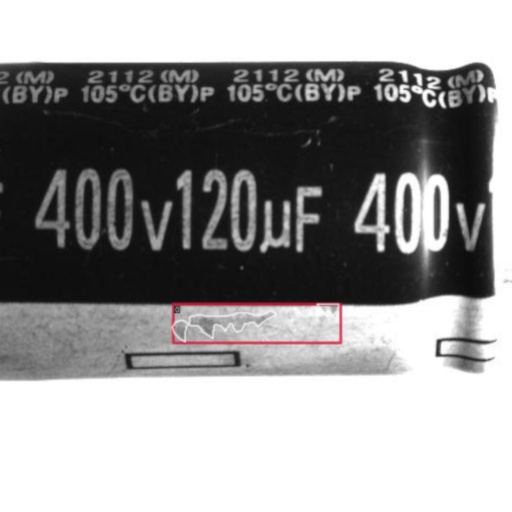}
    \caption*{(2)~Capacitor}
  \end{subfigure}
  \hfill
  \begin{subfigure}{0.138\linewidth}
    \includegraphics[width=0.98\linewidth]{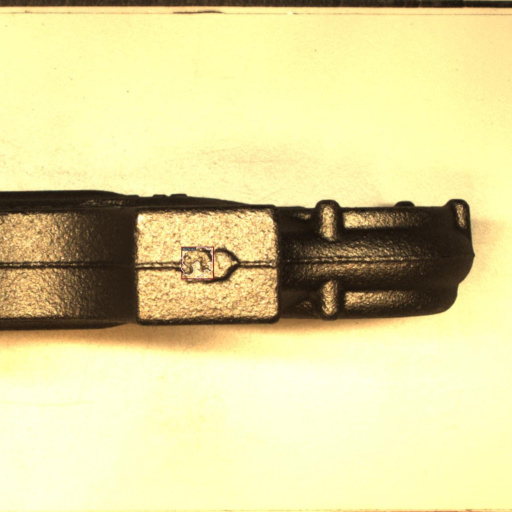}
    \caption*{(3)~Casting}
  \end{subfigure}
  \hfill
  \begin{subfigure}{0.138\linewidth}
    \includegraphics[width=0.98\linewidth]{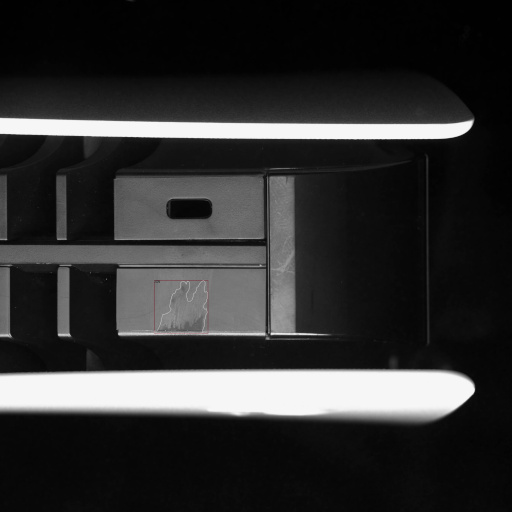}
    \caption*{(4)~Console}
  \end{subfigure}
  \hfill
  \begin{subfigure}{0.138\linewidth}
    \includegraphics[width=0.98\linewidth]{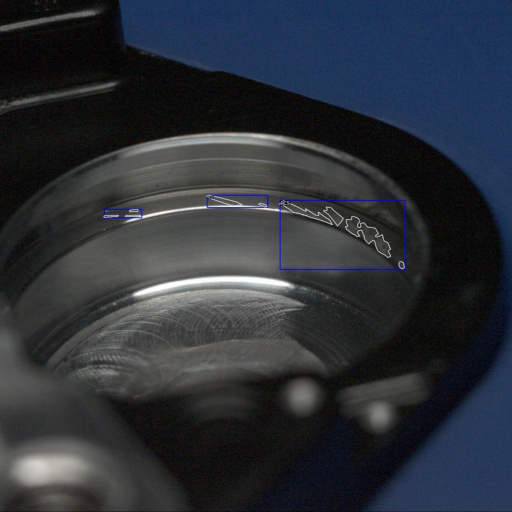}
    \caption*{(5)~Cylinder}
  \end{subfigure}
  \hfill
  \begin{subfigure}{0.138\linewidth}
    \includegraphics[width=0.98\linewidth]{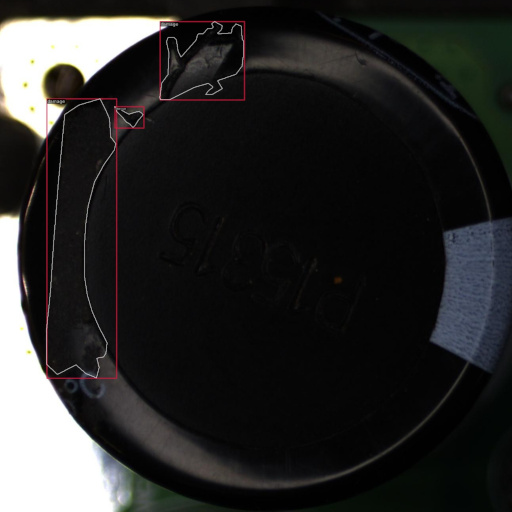}
    \caption*{(6)~Electronics}
  \end{subfigure}
  \hfill
  \begin{subfigure}{0.138\linewidth}
    \includegraphics[width=0.98\linewidth]{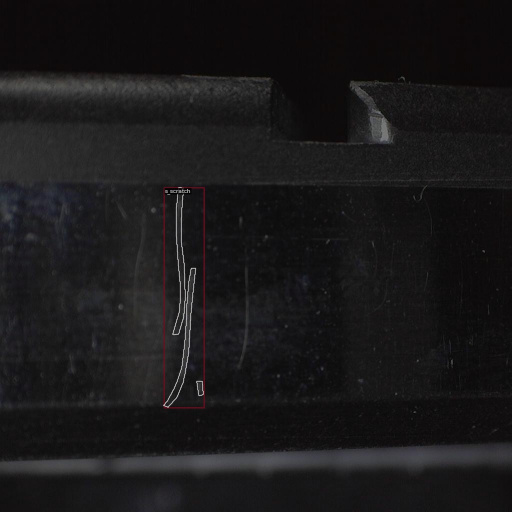}
    \caption*{(7)~Groove}
  \end{subfigure}
  \\
  \begin{subfigure}{0.138\linewidth}
    \includegraphics[width=0.98\linewidth]{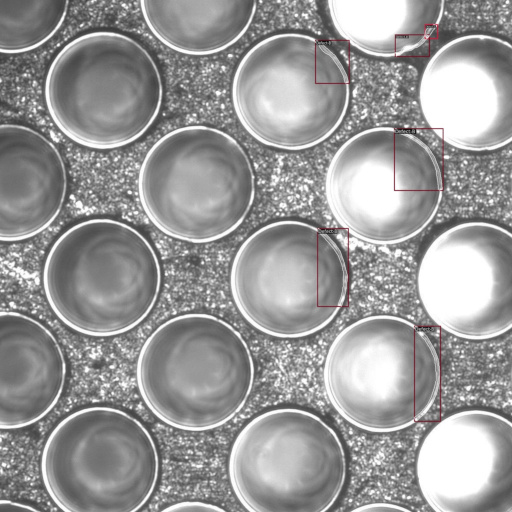}
    \caption*{(8)~Hemisphere}
  \end{subfigure}
  \hfill
  \begin{subfigure}{0.138\linewidth}
    \includegraphics[width=0.98\linewidth]{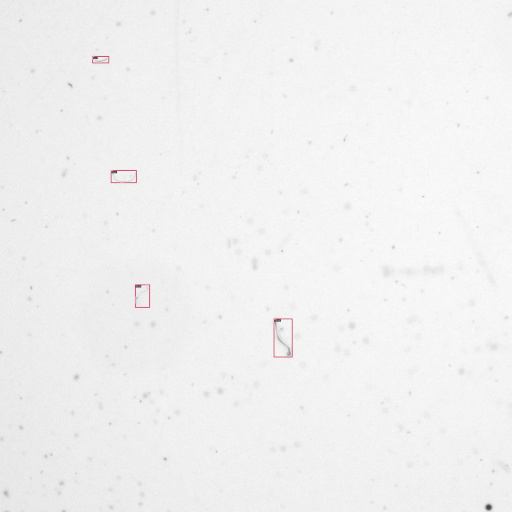}
    \caption*{(9)~Lens}
  \end{subfigure}
  \hfill
  \begin{subfigure}{0.138\linewidth}
    \includegraphics[width=0.98\linewidth]{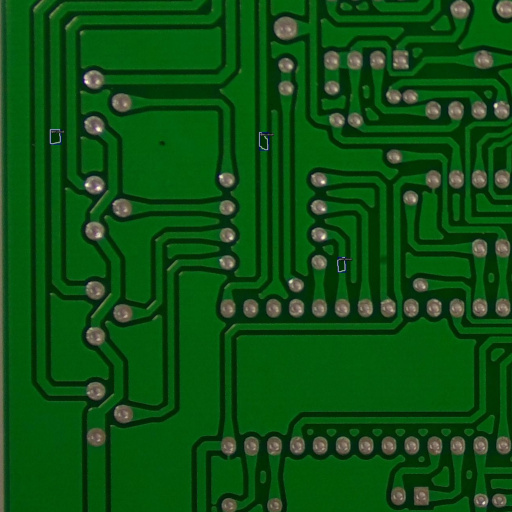}
    \caption*{(10)~PCB\_1}
  \end{subfigure}
  \hfill
  \begin{subfigure}{0.138\linewidth}
    \includegraphics[width=0.98\linewidth]{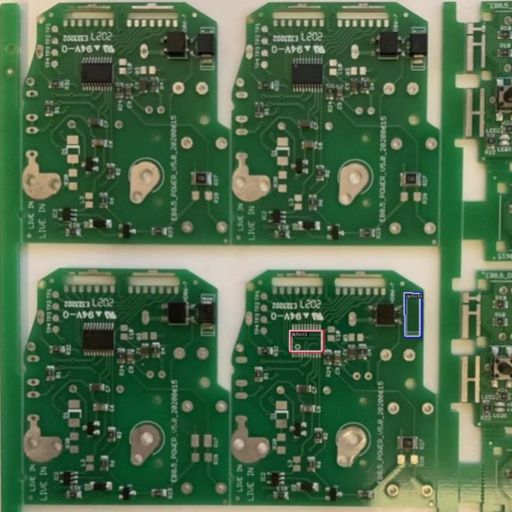}
    \caption*{(11)~PCB\_2}
  \end{subfigure}
  \hfill
  \begin{subfigure}{0.138\linewidth}
    \includegraphics[width=0.98\linewidth]{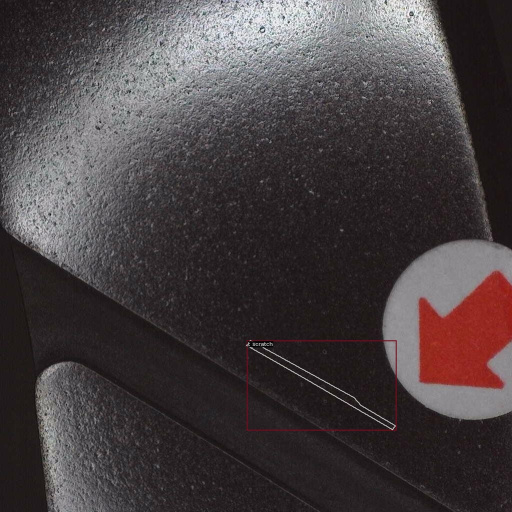}
    \caption*{(12)~Ring}
  \end{subfigure}
  \hfill
  \begin{subfigure}{0.138\linewidth}
    \includegraphics[width=0.98\linewidth]{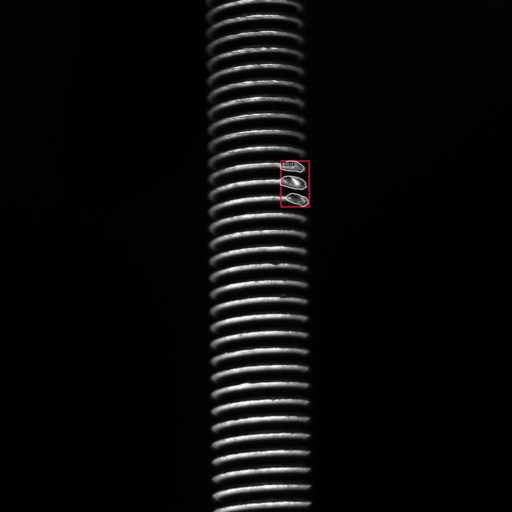}
    \caption*{(13)~Screw}
  \end{subfigure}
  \hfill
  \begin{subfigure}{0.138\linewidth}
    \includegraphics[width=0.98\linewidth]{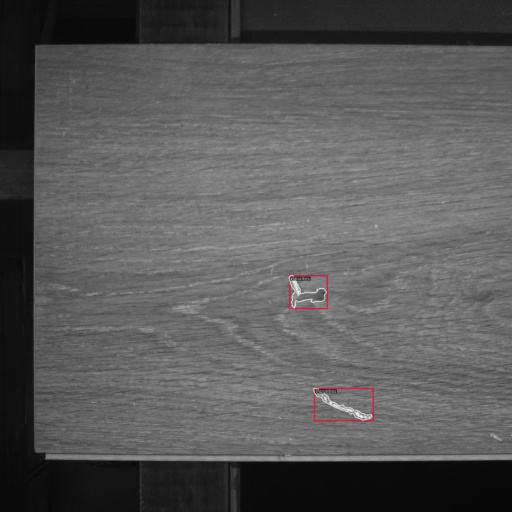}
    \caption*{(14)~Wood}
  \end{subfigure}

  \caption{Typical sample in each dataset used in this challenge. The images shown here have been cropped to a fixed size for display purposes, but the sizes of the raw images vary across different datasets.}
  \label{fig:dataset_example}
\end{figure*}

The CVPR 2023 Vision Challenge Track 1 for Data-Efficient Defect Detection is providing 14 datasets from \url{roboflow.com} as the official dataset. These datasets encompass a wide range of manufacturing processes, materials, and industries, with typical samples visualized in \cref{fig:dataset_example}. The defects present in these datasets differ significantly in terms of type, scale, and texture background. While Capacitor, Casting, Groove, Hemisphere, Ring, Screw, and Wood datasets contain images of only one size, multiple sizes are present in other datasets, with some even changing from [262, 192] to [2624, 2624]. Concerning defect classification, Capacitor, Electronics and Screw have only one category of defect while the remaining datasets feature multiple categories of defects.

\subsection{Implementation details}

In our experiment, we employed the MMDetection~\cite{mmdetection} framework for the competition, and executed the corresponding program using two A100 GPU. The baseline used for single instance segmentation involved HTC without semantic branches, and its learning rate was set to 0.0001. We repeated the dataset five times and train all models for total 10 epochs by default, with learning rate divided by 10 at the 2nd, 6th and 9th epoch. All models were pre-trained with ImageNet-22k initialization. In order to mitigate overfitting in the event of small samples, we froze only the first two layers on the backbone. To improve recall, we use SoftNMS~\cite{bodla2017soft} rather than NMS for the detector post processing. The training and validation sets were leveraged for network model training.

To further enhance the performance of instance segmentation, we adopted multi-scale training and TTA. Randomly scaling the image to any size from a predetermined list [(1024, 1024), (1536, 1536), (2048, 2048), (2560, 2560), (3072, 3072), (3584, 3584), (4096,4096)] and uniformly cropping it to a fixed size of (2048, 2048) was used for training purposes. During the testing phase, images were inferred based on a consistent multi-scale list [(1024, 1024), (1536, 1536), (2048, 2048), (2560, 2560), (3072, 3072), (3584, 3584), (4096, 4096)].

\begin{table*}[h]
  \centering
  \begin{tabular}{p{1.5cm} p{0.3cm} p{0.42cm} p{0.42cm} p{0.42cm} p{0.42cm} p{0.42cm} p{0.42cm} p{0.42cm} p{0.42cm} p{0.42cm} p{0.42cm} p{0.42cm} p{0.42cm} p{0.42cm} p{0.42cm} p{0.45cm} p{1.5cm}}
    \toprule
    ~ & ~ & \rot{Cable} & \rot{Capacitor} & \rot{Casting} & \rot{Console} & \rot{Cylinder} & \rot{Electronics} & \rot{Groove} & \rot{Hemisphere} & \rot{Lens} & \rot{PCB\_1} & \rot{PCB\_2} & \rot{Ring} & \rot{Screw} & \rot{Wood} & \rot{\small \bf Average} & \rot{\footnotesize \bf (mAP+mAR)/2} \\
    \midrule
    \textbf{HTC} &{\small \it mAP}& 42.7 & 54.4 & 28.2 & 27.7 & 59.1 & 36.2 & 30.2 & 44.0 & 43.8 & 81.5 & 93.9 & 23.0 & 58.3 & 38.2 & 47.20 & \multirow{2}{*}{\bf ~~~~54.17} \\
               ~ &{\small \it mAR}& 56.1 & 69.8 & 45.2 & 44.4 & 72.1 & 55.9 & 54.1 & 58.0 & 56.9 & 86.9 & 96.1 & 39.0 & 69.4 & 52.1 & 61.14 & \\
    \midrule
    \textbf{R50} &{\small \it mAP}& 31.6 & 41.5 & 21.4 & 10.7 & 54.4 & 21.6 & 23.0 & 40.0 & 38.6 & 80.3 & 92.7 & 18.5 & 55.7 & 32.4 & 40.17 & \multirow{2}{*}{\bf ~~~~45.10} \\
               ~ &{\small \it mAR}& 41.4 & 52.1 & 34.5 & 23.4 & 63.7 & 32.6 & 37.5 & 55.1 & 50.0 & 85.1 & 94.9 & 26.4 & 62.8 & 40.9 & 50.03 & \\
    \midrule
    \textbf{ConvNext} &{\small \it mAP}& 36.2 & 24.7 & 22.2 & 23.7 & 56.0 & 32.4 & 28.8 & 39.0 & 44.2 & 80.3 & 93.3 & 18.2 & 54.8 & 32.3 & 41.86 & \multirow{2}{*}{\bf ~~~~47.36} \\
                    ~ &{\small \it mAR}& 49.4 & 39.0 & 31.1 & 34.2 & 65.4 & 48.1 & 43.0 & 54.1 & 56.1 & 85.6 & 95.6 & 31.6 & 64.7 & 42.0 & 52.85 & \\
    \bottomrule
  \end{tabular}
  \caption{Results of different instance segmentation models. HTC, R50, and ConvNext are shortened for HTC-SwinB-CBNetV2, Cascade-Mask-RCNN-ResNet50, and Cascade-Mask-RCNN-ConvNext, respectively.}
  \label{tab:instance_result}
\end{table*}

\begin{table*}[h]
  \centering
  \begin{tabular}{p{2.4cm} p{0.3cm} p{0.40cm} p{0.40cm} p{0.40cm} p{0.40cm} p{0.40cm} p{0.40cm} p{0.40cm} p{0.40cm} p{0.40cm} p{0.40cm} p{0.40cm} p{0.40cm} p{0.40cm} p{0.40cm} p{0.45cm} p{1.5cm}}
    \toprule
    ~ & ~ & \rot{Cable} & \rot{Capacitor} & \rot{Casting} & \rot{Console} & \rot{Cylinder} & \rot{Electronics} & \rot{Groove} & \rot{Hemisphere} & \rot{Lens} & \rot{PCB\_1} & \rot{PCB\_2} & \rot{Ring} & \rot{Screw} & \rot{Wood} & \rot{\small \bf Average} & \rot{\footnotesize \bf (mAP+mAR)/2} \\
    \midrule
    \textbf{HTC} &{\small \it mAP}& 42.7 & 54.4 & 28.2 & 27.7 & 59.1 & 36.2 & 30.2 & 44.0 & 43.8 & 81.5 & 93.9 & 23.0 & 58.3 & 38.2 & 47.20 & \multirow{2}{*}{\bf ~~~~54.17} \\
               ~ &{\small \it mAR}& 56.1 & 69.8 & 45.2 & 44.4 & 72.1 & 55.9 & 54.1 & 58.0 & 56.9 & 86.9 & 96.1 & 39.0 & 69.4 & 52.1 & 61.14 & \\
    \midrule
    \textbf{HTC} {\bf \small +segm} &{\small \it mAP}& 42.7 & 54.4 & 28.2 & 29.1 & 62.9 & 38.4 & 30.2 & 44.7 & 51.0 & 81.5 & 93.9 & 23.2 & 59.0 & 38.2 & 48.38 & \multirow{2}{*}{\bf ~~~~55.41} \\
                               ~    &{\small \it mAR}& 56.8 & 69.8 & 46.0 & 46.2 & 73.4 & 57.3 & 54.7 & 58.0 & 63.0 & 86.9 & 96.1 & 42.2 & 71.0 & 52.7 & 62.43 & \\
    \midrule
    \textbf{HTC} {\bf \footnotesize +segm +R50}  &{\small \it mAP}& 42.7 & 54.4 & 28.2 & 29.1 & 62.9 & 38.4 & 30.2 & 44.7 & 51.0 & 81.5 & 93.9 & 23.2 & 59.0 & 38.2 & 48.38 & \multirow{2}{*}{\bf ~~~~55.94} \\
                                            ~    &{\small \it mAR}& 57.8 & 70.9 & 47.0 & 46.2 & 74.4 & 58.4 & 55.7 & 62.0 & 64.0 & 87.9 & 96.5 & 44.0 & 71.2 & 52.9 & 63.49 & \\
    \midrule
    \textbf{HTC} {\bf \footnotesize +segm +R50}  &{\small \it mAP}& 42.8 & 54.4 & 28.2 & 29.1 & 63.4 & 38.4 & 30.2 & 44.9 & 51.2 & 81.6 & 93.9 & 23.5 & 59.0 & 38.2 & 48.49 & \multirow{2}{*}{\bf ~~~~57.60} \\
      {\bf \footnotesize ~~~~~~~~~~~~+ConvNext} &{\small \it mAR}& 63.0 & 75.2 & 49.8 & 49.2 & 78.7 & 59.5 & 58.5 & 71.0 & 68.5 & 90.7 & 97.5 & 45.6 & 71.7 & 54.6 & 66.71 & \\

    \bottomrule
  \end{tabular}
  \caption{Results of fusing different models. HTC, segm, R50, and ConvNext are shortened for HTC-SwinB-CBNetV2, Mask2Former, Cascade-Mask-RCNN-ResNet50, and Cascade-Mask-RCNN-ConvNext, respectively.}
  \label{tab:fuse_result}
\end{table*}

\subsection{Ablation study}

\subsubsection{Comparison of instance segmentation models}
\ 
\indent In this section, we compare the different instance segmentation models. All the models are evaluated in the test dataset. The results are shown in \cref{tab:instance_result}. 

\subsubsection{Model ensemble result}
\ 
\indent Model ensemble is a classic and highly effective technique for ensuring robust performance. Our approach employs a set of models known as model bags, which comprises one semantic segmentation model Mask2Former, and two instance segmentation models, namely Cascade-Mask-RCNN-ResNet50 and Cascade-Mask-RCNN-ConvNext. These models were purposefully designed to increase diversity between them.

Mask2Former is a sophisticated segmentation architecture that has been proven effective in delivering state-of-the-art results in various applications such as semantic, instance, and panoramic segmentation. By fusing semantic segmentation with instance segmentation, we achieve a remarkable mask mAP of 48.38\% on the test dataset (as shown in \cref{tab:fuse_result}). Finally, by averaging the predictions of these models within the model bags, our model ensemble delivers outstanding performance in the competition, with a mAP of 48.49\% and a mAR of 66.71\%.

\section{Potential Improvements}
\label{sec:improv}
\textbf{Semi-supervised learning}: In our experiments, we focused on training instance segmentation models solely on the training and validation sets. We try to use the semi supervised learning method based on soft-teacher~\cite{xu2021end} to improve the performance of instance segmentation. However, we found that due to differences in the dataset, it is not possible to provide a unified training strategy for semi-supervised models. Due to constraints on competition time, employing semi-supervised methods as a future research direction seems more feasible.

\textbf{SAM}: Meta proposed Segment Anything Model (SAM)~\cite{kirillov2023segment} as a fundamental model for addressing segmentation tasks. We evaluated its effectiveness through an online demo website and determined that this model has excellent generalization performance on industrial data as well. However, given the competition rules, we cannot use the SAM. Nonetheless, large or fundamental models have the potential to bring significant changes to industrial defect detection, thus providing another direction for improvement in future work.

\section{Conclusion}
\label{sec:con}
In this report, we present the technical details of our submission to the CVPR 2023 Vision Challenge Track 1 for Data-Efficient Defect Detection challenge. Our approach consists of three main components: a base instance segmentation model, a method that integrates semantic segmentation into instance segmentation, and a strategy that fuses multiple instance segmentations. Through a series of experiments, we demonstrate the competitiveness of our approach on the test set, achieving an average mAP@0.50:0.95 of more than 48.49\% and an average mAR@0.50:0.95 of 66.71\%.

\section*{Acknowledgment}
We would like to express our gratitude to Xilong Liu and Shuhan Shen from Institute of Automation, Chinese Academy of Sciences for providing GPUs for our model training.

{\small
\bibliographystyle{ieee_fullname}
\bibliography{egbib}
}

\end{document}